\documentclass[conference]{IEEEtran}
\IEEEoverridecommandlockouts
\usepackage{cite}
\usepackage{amssymb,amsfonts}
\usepackage{amsmath}
\usepackage{algorithmic}
\usepackage{textcomp}
\usepackage{graphicx}
\usepackage{lipsum} 
\usepackage{mwe}
\usepackage{caption}    
\usepackage[lofdepth,lotdepth]{subfig}
\usepackage{caption}
\usepackage{scrextend}
\usepackage{multirow}
\usepackage[para,online,flushleft]{threeparttable}

\newcommand{\w}{{\mathbf{w}}}
\newcommand{\x}{{\mathbf{x}}}
\newcommand{\y}{{\mathbf{y}}}  
\newcommand{\tb}{{\mathbf{{t}}}}
\newcommand{\ub}{{\mathbf{{u}}}}
\newcommand{\vb}{{\mathbf{{v}}}}
\newcommand{\R}{{\mathbb{R}}}

\addtokomafont{labelinglabel}{\sffamily}

\def\BibTeX{{\rm B\kern-.05em{\sc i\kern-.025em b}\kern-.08em
    T\kern-.1667em\lower.7ex\hbox{E}\kern-.125emX}}
\begin{document}

\title{Elastic Neural Networks: A Scalable Framework for Embedded Computer Vision\\
\thanks{This work was partially funded by the Academy of Finland
project 309903 CoefNet, and by TEKES --- the Finnish Technology Agency for Innovation (FiDiPro project StreamPro 1846/31/2014). Authors also thank CSC--IT Center for Science for computational resources.
}}

\author{\IEEEauthorblockN{Yue Bai} 
\IEEEauthorblockA{
\textit{Tampere Univ. of Tech.}\\
Tampere, Finland}

\and \IEEEauthorblockN{Shuvra S. Bhattacharyya} \IEEEauthorblockA{
\textit{University of Maryland, USA, and} \\ \textit{Tampere Univ. of Tech., Finland}}
\and 

\IEEEauthorblockN{Antti P. Happonen} \IEEEauthorblockA{
\textit{Tampere Univ. of Tech.}\\
Tampere, Finland}

\and \IEEEauthorblockN{Heikki Huttunen}
\IEEEauthorblockA{
\textit{Tampere Univ. of Tech.}\\
Tampere, Finland}
}
\maketitle

\begin{abstract}
We propose a new framework for image classification with deep neural networks. The framework introduces intermediate outputs to the computational graph of a network. This enables flexible control of the computational load and balances the tradeoff between accuracy and execution time. 

Moreover, we present an interesting finding that the intermediate outputs can act as a regularizer at training time, improving the prediction accuracy. In the experimental section we demonstrate the performance of our proposed framework with various commonly used pretrained deep networks in the use case of apparent age estimation.
\end{abstract}


\section{Introduction}

Deep learning has rapidly become the state of the art in modern machine learning, and has surpassed the human level in several classical problems. Majority of research concentrates on the accuracy of the model, but also the computational load has been studied. This line of research attempts to reduce the execution time of the forward pass of the network. Well-known techniques for constructing lightweight yet accurate networks include decomposing the convolutions either in horizontal and vertical dimensions \cite{alvarez2016decomposeme}, or in spatial and channel dimensions \cite{howard2017mobilenets}, or using a fully convolutional architecture
\cite{iandola2016squeezenet}.

In non-static environments, the computational requirements may have a large variation. For example, in the use case of this paper---real-time age estimation---the number of faces seen by the camera directly influences the amount of computation. Namely, each detected face is sent to the GPU for age assessment. Thus, the workload for 10 faces in front of the camera is 10-fold compared to just a single face. Additionally, the same network should be deployable over a number of architectures: For example, in 2015 there were already over 24,000 different hardware architectures running the Android operating system. Thus, there is a demand for an \textit{elastic network structure}, able to adjust the computational complexity on the fly. The network should be monolithic (as opposed to just training several networks with different complexities), since the storage and computational overhead of deploying a large number of separate networks would be prohibitive in many applications.

Recently, Huang \textit{et al.} \cite{huang2017multi} proposed a network architecture addressing these problems. Their \textit{Multi-Scale Dense Network} (MSDNet) adds early-exits to the computational graph, and stores a representation of the input at several resolutions throughout the network. This way, the computational time can be adjusted even to the extent of \textit{anytime prediction}, where the network almost always has some prediction at hand.

In this work, we take a simpler approach: Instead of proposing a new network topology, we study the effect of adding early-exits to \textit{any network}. Namely, there exist a number of successful network topologies pretrained with large image databases and the progress of deep learning is so rapid, that new network topologies appear daily. We propose a general framework that systematically ``elastifies'' an arbitrary network to provide novel trade-offs between execution time and accuracy, while leveraging key properties of the original network.


\begin{figure*}[t]
\vspace{1mm}
\centering
\includegraphics[width=0.9\textwidth]{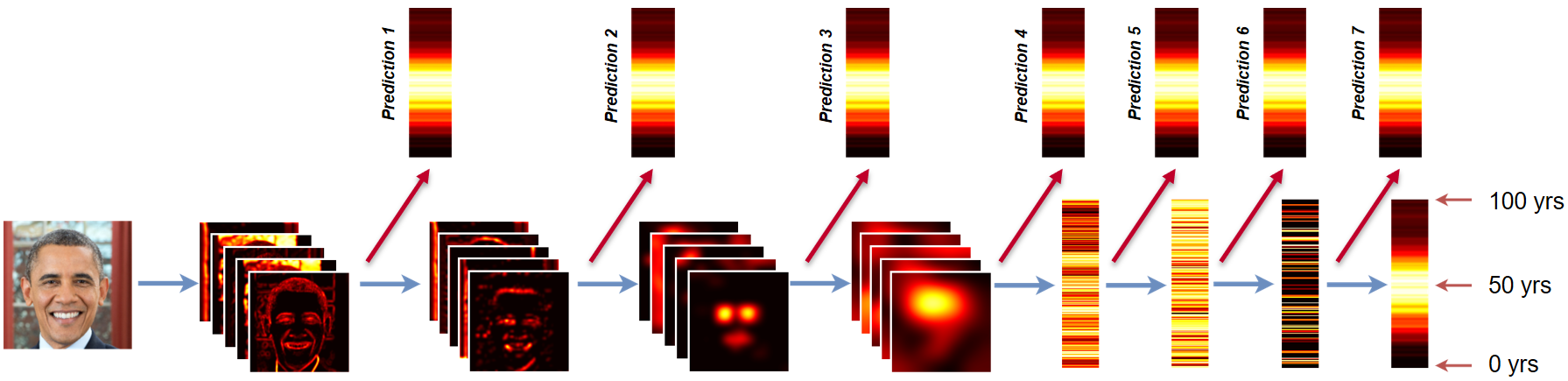}
\caption{The proposed neural network structure with early exits.}
\label{fig:elastic}
\end{figure*}

\section{Elastic Deep Neural Networks}
\label{sec:elastic}





\begin{figure}%
    \centering
    \subfloat[]{{\includegraphics[width=0.45\linewidth]{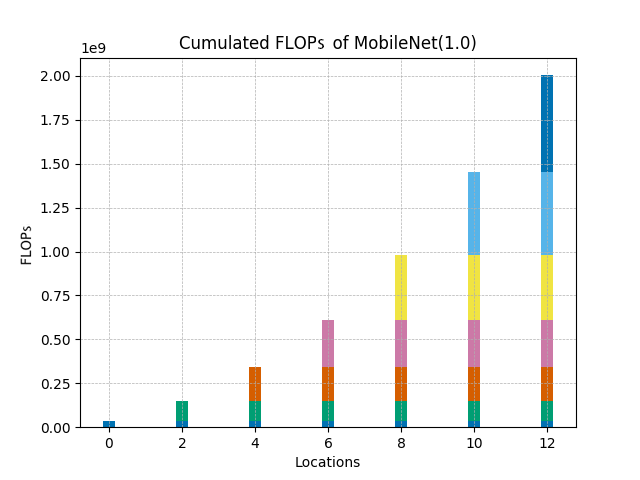} }}%
    \label{fig:mobile}
    \subfloat[]{{\includegraphics[width=0.45\linewidth]{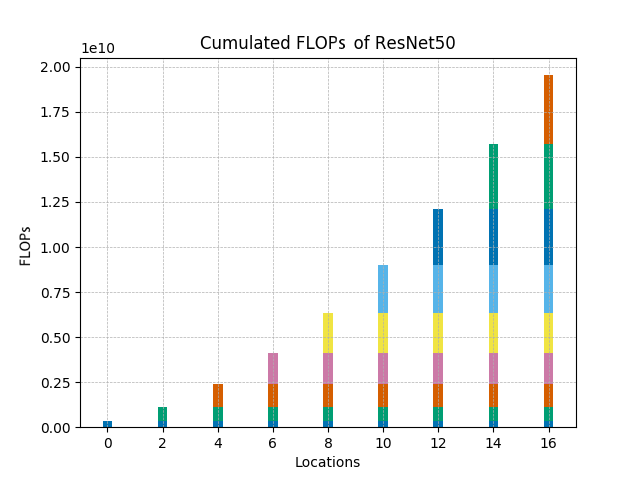} }}%
    \label{fig:res}
    \caption{Cumulative FLOPs on MobileNet ($\alpha$ = 1) and ResNet50.}%
    \label{fig:cum-flops}%
    
\end{figure}

In deep neural networks, the time to complete the forward (prediction) pass is composed of the execution times of individual layers. As an example, Figure \ref{fig:cum-flops} shows the cumulative floating-point operations (FLOPs) per image frame (i.e., for each iteration of the neural network) for different state-of-the-art network structures. Conventionally, the network outputs the class probabilities of the input image at the final classifier layer, thus summing up the number of FLOPs spent at the early layers. The final classifier layer is also the most accurate one to categorize the input.
However, the feature maps of the earlier layers also possess some predictive power, leading to the question of whether the early layer features are useful for prediction, as well.
Thus, we introduce a classifier structure with early-exits, which base their prediction on the features extracted at intermediate layers. An illustration of the proposed structure is shown in Figure \ref{fig:elastic}. The figure illustrates the elastic structure in an apparent age estimation context, where the input RGB image has dimensions $224\times 224 \times 3$ and the prediction target is a vector of probabilities for ages $y \in \{0,1,2,\ldots, 100\}$. In particular, the figure shows the early exits, which are labeled Prediction 1, Prediction 2, ..., Prediction 7.

For each early exit (and the final one), we set exactly the same training target and loss function. More specifically, our loss function for a training sample $(\x, \y)$, with input $\x\in \R^{w\times h \times d}$ and one-hot-encoded class label $\y\in \R^C$ (with $C$ classes), is defined as:
\begin{equation}
L(\w; \x,\y) = \sum_{p=1}^{P} \alpha_p \ell(\y, \hat{\y}_p(\x; \w)),
\end{equation}
where $\hat{\y}_p(\x; \w)$ denotes the $p$'th output in the pipeline ($p=1,2,\ldots, P$) for network with parameters $\w$ and input $\x$; $\alpha_p >0$ is the weight for each output; and $\ell(\cdot, \cdot)$ denotes a loss function (\textit{e.g.,} categorical cross-entropy).

In practice, the number of layers may be high (even hundreds of layers), and the number of early exits may grow unnecessarily high and slow down the training process if an early exit is inserted after every layer. However, most modern networks are constructed of repeated blocks, where each block contains multiple layers. By placing early exits only at the block outputs, we can significantly reduce the burden on the training process. Moreover, the very early blocks are not useful for prediction, so we assign zero weights $\alpha_p = 0$ to those early exit losses.

%

%


\section{Apparent Age Estimation}
\label{sec:age_estimation}

\begin{figure}[t]
\begin{center}
\includegraphics[width=0.40\columnwidth]{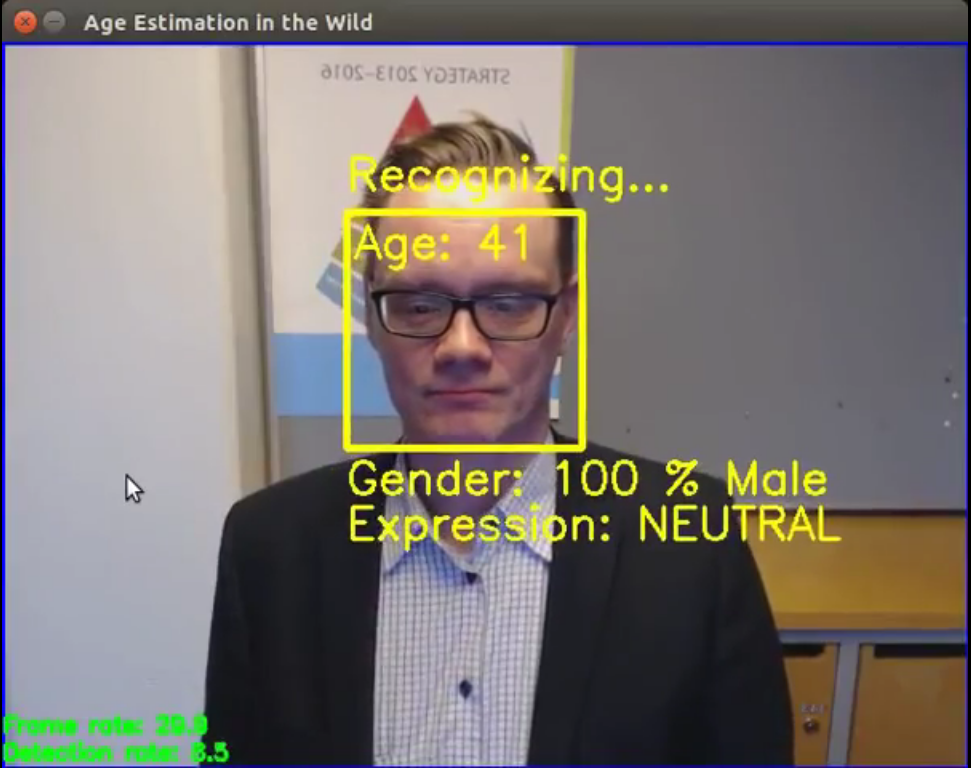}
\end{center}
\caption{Real time age, gender and facial expression recognizer.}
\label{fig:use-case}%
\end{figure}

We investigate the performance of the proposed elastic networks methodology in real-time apparent age estimation. Figure \ref{fig:use-case} shows an example of the application which is related to our real time age estimation demo
\cite{HuttunenHandbook}, where people may freely enter in front of a screen and a camera, and the system then detects all faces and assesses the age, gender and facial expression of each based on the cropped bounding box of each face. When multiple people appear in front of the screen, the system has to evaluate the ages for each face separately, and the computational load increases accordingly. Thus, the need for flexible control of the computational load is evident.

Let us now describe the detailed training protocol: datasets, preprocessing, and the network structures used. 

\subsection{Datasets}

Following the procedure of \cite{rothe2015dex}, we use altogether three datasets for training. First, we initialize our network with Imagenet \cite{imagenet_cvpr09} pretrained weights, then we pretrain the network for faces using the IMDB-WIKI\cite{rothe2015dex} facial database (large but noisy), and finally  train the network using a smaller human annotated ChaLearn LAP dataset from CVPR2016 competition (small but clean).

\textbf{\em IMDB-WIKI}---The IMDB-WIKI dataset is the largest publicly available dataset of facial images with gender and age labels for training. The facial images in this dataset are crawled from the IMDB and Wikipedia websites. There are 461,871 facial images from the IMDB website and 62,359 from the Wikipedia website. At each epoch, we feed the network with 32,768 randomly chosen images with left-right-flip augmentation and iterate for 100 epochs.

\textbf{\em CVPR 2016 ChaLearn Looking at People Competition on apparent age estimation dataset}---%
In the apparent age estimation competition organized at IEEE CVPR 2016 \cite{escalera2016chalearn}, a state-of-the-art dataset was released with altogether 7,591 facial images (4,113 in the training dataset, 1,500 in the validation dataset and 1,978 in the testing dataset) with human-annotated apparent ages and standard deviations. 
Compared to other facial age datasets, the images in this dataset are taken in non-controlled environments and have diverse backgrounds. 

\subsection{Image Preprocessing}

\textbf{\em Face detection}---%
During the real-time prediction, we use the Viola-Jones \cite{viola2001rapid} face detector. Despite its poor accuracy compared to modern CNN detectors, it is still among the fastest algorithms when not using the GPU (which we wish to reserve for the age estimation). The deficiencies in accuracy are compensated by the streaming nature of the data source; not every frame has to be detected. 

At training time, however, we want to exploit every training image---even those at poses difficult for the Viola-Jones detector. Therefore, We use the open source `Head Hunter' \cite{Mathias2014Eccv} face detector to detect the bounding boxes. 
If there are multiple faces are detected in one image, we will choose the first one or the one with the highest score.
Note that the discrepancy of using a different detector at training and testing time is compensated by the subsequent alignment step, where we align the face to a template set of facial landmarks.


\textbf{\em Landmark Detection}---%
Alignment of the face to fixed coordinates will greatly improve the prediction accuracy. To this aim, we will first find the landmarks for each face found by the detector. In our implementation, we use the regression forest method of Kazemi \textit{et al.} \cite{kazemi2014one}, due to its fast speed. In our landmark set, we use the \textit{dlib}\footnote{http://dlib.net/} implementation with 68 landmarks, of which we use 47 central keypoints, discarding the boundary keypoints often occluded due to pose. We match the found facial keypoints with a set of target keypoints by using a similarity transformation matrix.

\begin{figure}[t]
\centering
\subfloat[Subfigure 1 list of figures text][]{
\includegraphics[width=0.1\textwidth]{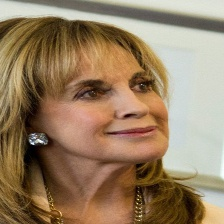}
\label{fig:subfig1}}
\subfloat[Subfigure 2 list of figures text][]{
\includegraphics[width=0.1\textwidth]{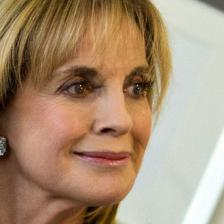}
\label{fig:subfig2}}
\subfloat[Subfigure 3 list of figures text][]{
\includegraphics[width=0.1\textwidth]{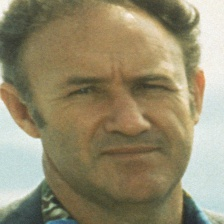}
\label{fig:subfig3}}
\subfloat[Subfigure 4 list of figures text][]{
\includegraphics[width=0.1\textwidth]{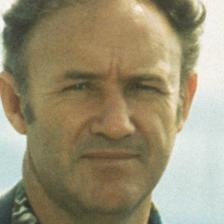}
\label{fig:subfig4}}
\caption{The result of alignment with full affine transformation is shown in (a) and (c), while (b) and (d) depicts the similarity transformation without shearing.}
\label{fig:globfig}
\end{figure}

\begin{figure}
\centering
\includegraphics[width=0.45\columnwidth]{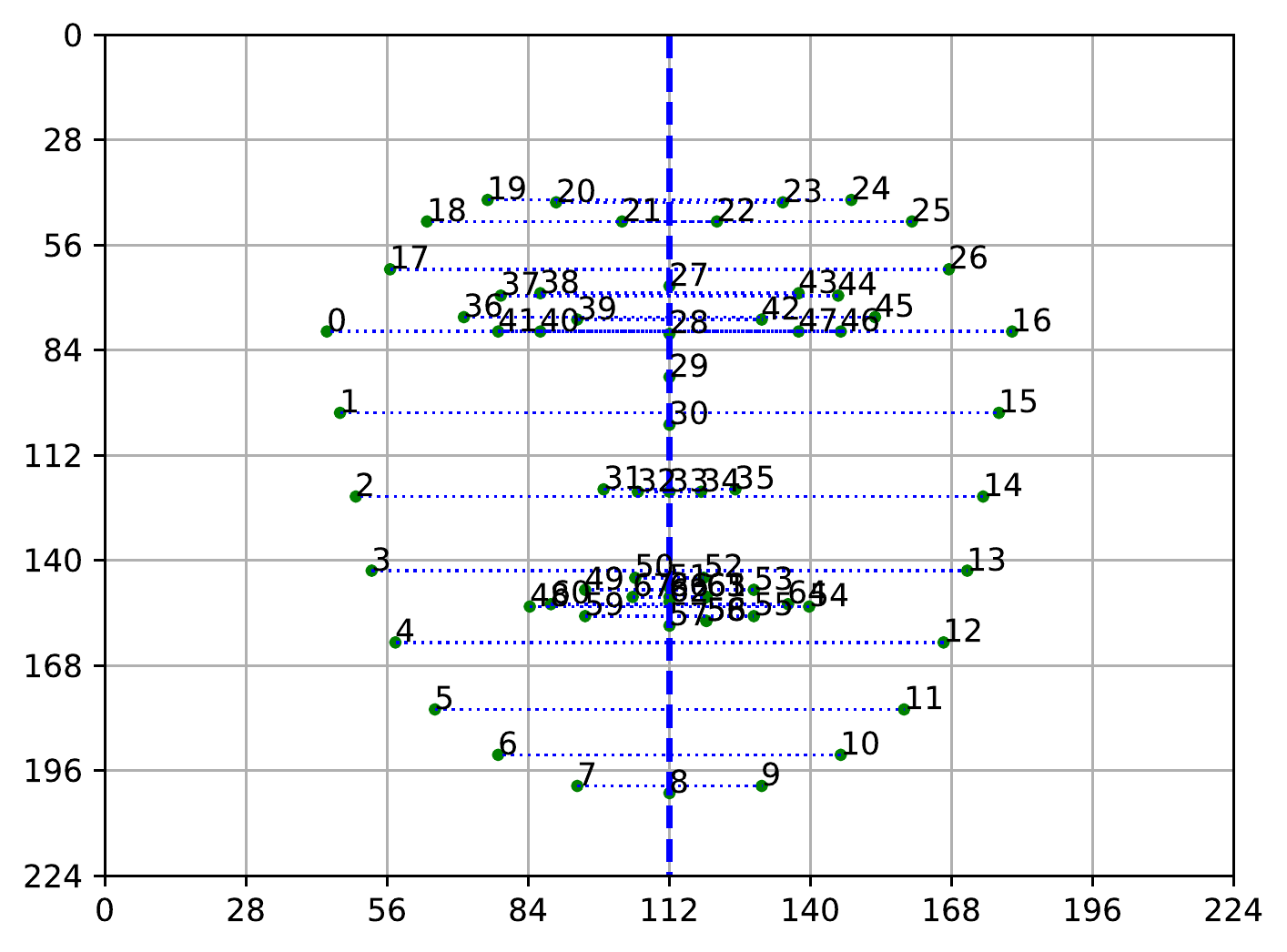}
\caption{Landmark targets for the alignment. Symmetric landmarks pairs are highlighted by dashed lines.}\label{fig:landmarks}
\end{figure}

\textbf{\em Target Landmarks}---%
The landmark template is obtained from the landmarks of a sample face from the dataset. However, we normalize the template such that the landmarks are horizontally symmetric with respect to the centerline of the face. This is done in order to allow training set augmentation by adding horizontal flips of each training face. More specifically, we manually marked symmetric pairs of landmarks, and averaged their vertical coordinates and distances from the horizontal center location. Finally, the resulting set of coordinates was scaled to fit the network input size of $224\times 224$ pixels leaving 10\% margin at the bottom edge and 20\% margin at the other edges. 
The resulting landmark template is illustrated in Figure~\ref{fig:landmarks}.

\textbf{\em Face Alignment}---The simple approach of using the full affine transformation with least squares fit will distort the image and degrade the estimation performance. This is due to the shearing component as shown in Figure \ref{fig:subfig1} and Figure \ref{fig:subfig3}. This way the shape of faces in original images can be affected badly, which we wish to avoid. Instead, we use the similarity transformation allowing only rotation, scale and translation. The definition of the similarity transformation mapping points $\ub\in\R^2 \mapsto \vb\in\R^2$ with translation $\tb=(t_x,t_y)^T$, scaling $s\in \R_+$ and rotation angle $\theta$ is:
\begin{equation}
\vb=\boldsymbol{H}_{S}\x=
\begin{bmatrix}
    s\boldsymbol{R}       & \tb   \\
    \boldsymbol{0}^{T}      & 1   \\
\end{bmatrix}\ub =
\begin{bmatrix}
    s\cos\theta    & -s\sin\theta & t_{x}  \\
    s\sin\theta      & s\cos\theta   & t_{y}\\
    0      & 0   & 1\\
\end{bmatrix} \ub  \label{eqn.similarity}
\end{equation}

According to Hartley and Zisserman \cite{hartley2003multiple}, the solution can be derived from the vector cross product of point correspondences in homogeneous coordinates, $\ub_i=(x_i,y_i,1)^{T}$ and $\vb_i=(x'_i,y'_i,1)^{T}$, as 
\begin{equation}
\vb_i\times\boldsymbol{H}_{S}\ub_i=0.
\label{eqn.dlt}
\end{equation}
Substituting Eq. (\ref{eqn.similarity}) into Eq. (\ref{eqn.dlt}), the system simplifies to \cite{KamPaa:2009}
\begin{equation}
\begin{bmatrix}
    -y_i   & -x_i & 0 & 1 \\
    x_i      & -y_i   & 1& 0\\
\end{bmatrix}
\begin{bmatrix}
    s\cos\theta  \\
    s\sin\theta \\
    t_x \\
    t_y \\
\end{bmatrix}=
\begin{bmatrix}
    -y'_i \\
    x'_i\\
\end{bmatrix},
\end{equation}
which can be solved by the singular value decomposition \cite{hartley2003multiple}. Finally, we construct the similarity matrix $\boldsymbol H_S$ by inserting the four variables into it. 
Examples of aligned pictures are shown as in Figure \ref{fig:subfig2} and Figure \ref{fig:subfig4}.

\subsection{Pre-Trained Networks}

We use three commonly used network structures, each pretrained with the 1000-class ImageNet dataset. For age estimation, we only take the convolutional pipeline of each and append a 101-class dense layer to the top (for 101 ages: $0,1,\ldots, 100$ years). In this section, we present three elastic network structures that we developed for apparent-age estimation, and we specify where the early exits are located in these networks.

\textbf{\em ResNet50 Based Elastic Neural Network}---%
ResNet50 \cite{he2016deep} has 50 convolution layers, composed mainly of 16 three-layer residual blocks. We insert early-exit classifiers after each residual block producing a total of 17 intermediate classifiers (plus the final classification layer) and set the weights nonzero starting at the midpoint of the pipeline: $\alpha_p = 0$ for $p < 9$ and $\alpha_p = 1$ for $p \ge 9$.

\textbf{\em MobileNet Based Elastic Neural Network}---%
MobileNets \cite{howard2017mobilenets} have altogether 28 convolutional layers, consisting of an input layer, 13 two-layer blocks (one depthwise and one pointwise convolution) and the output layers. We connect early-exit  classifiers to the outputs of all the 13 blocks and the final classification layer and set the weights nonzero starting at the midpoint of the pipeline: $\alpha_p = 0$ for $p < 7$ and $\alpha_p = 1$ for $p \ge 7$.

\textbf{\em Inception-V3 Based Elastic Neural Network}---%
Inception-V3 network \cite{szegedy2016rethinking} has altogether 94 convolutional layers (some in parallel) include 11 blocks. We insert early-exit classifiers after each blocks producing a total of 12 intermediate classifiers (plus the final classification layer) and set the weights nonzero starting at the midpoint of the pipeline: $\alpha_p = 0$ for $p < 6$ and $\alpha_p = 1$ for $p \ge 6$.

\section{Experimental Results}
\label{sec:experiments}

We implement our proposed models using Keras based on Tensor flow backend \cite{chollet2018deep}. All models are trained using stochastic gradient descent (SGD) with mini-batch size 32. We use momentum with a momentum weight of $0.9$ . 

The initialized weights for every output layer follow the uniform distribution. We give a weight for every output layer equal to one. During the pretraining phase, we first freeze all layers except output layers and train with a learning rate of $1 \times 10^{-3}$ for the first $ 10 $ epochs. Then we train all layers of all models for 150 epochs, with an initial learning rate of $1 \times 10^{-2}$, which is divided by a factor of 10 after the decaying loss on the validation set stays on a plateau. The minimum of learning rate is $1 \times10^{-5}$. We apply the same optimization scheme to the training phase, except that the initial learning rate is decreased to $1 \times 10^{-3}$.

\subsection{Evaluation Criteria}
We use two different criteria to evaluate the accuracy of age estimation. 

\textbf{$\mathit{MAE}$}---Mean Absolute Error (MAE) is the average  of absolute differences between the predicted and the real age:
\begin{equation}
\mathit{MAE} = \frac{1}{N}\sum_{n=1}^{N} |\hat{y}_n -y_n|, \label{eqn.mae}
\end{equation}
where $\hat{y}_n$ and $y_n$ are the estimated and ground-truth age of the $n$-th testing image, respectively. 

\textbf{\em $\epsilon$-error}---%
The degree of agreement among the human annotators  may be used as an indicator of difficulty of each image. In our training dataset, the minimum and maximum standard deviations are 0 and 12.4, while in the validation these are 0 and 14.1. To take the relative difficulty into account, the ChaLearn competition organizers defined an $\epsilon$-metric defined as
\[
\epsilon = 1 - \exp\left( -\frac{(y-\mu)^2}{2\sigma^2} \right),
\]
where $y$ is the age predicted by the algorithm, $\mu$ is the mean assessment of human evaluators, and $\sigma^2$ is their variance. We use it as another accuracy criterion in our experiments.


\subsection{Results}
\begin{figure*}%
    \centering
    \subfloat[$\epsilon$-error over different networks]{{\includegraphics[width=0.45\linewidth]{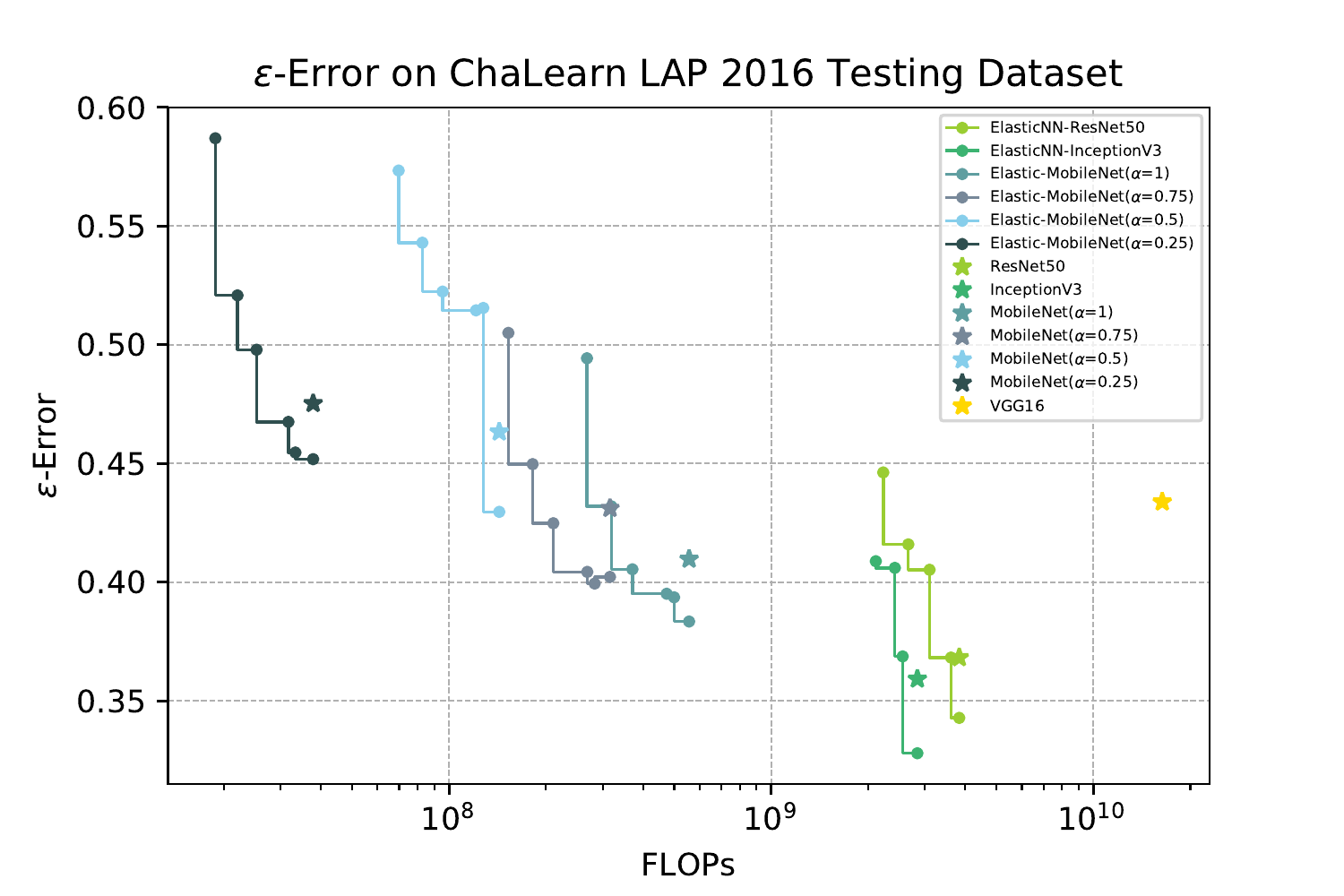} }}%
    \qquad
    \subfloat[MAE over different networks]{{\includegraphics[width=0.45\linewidth]{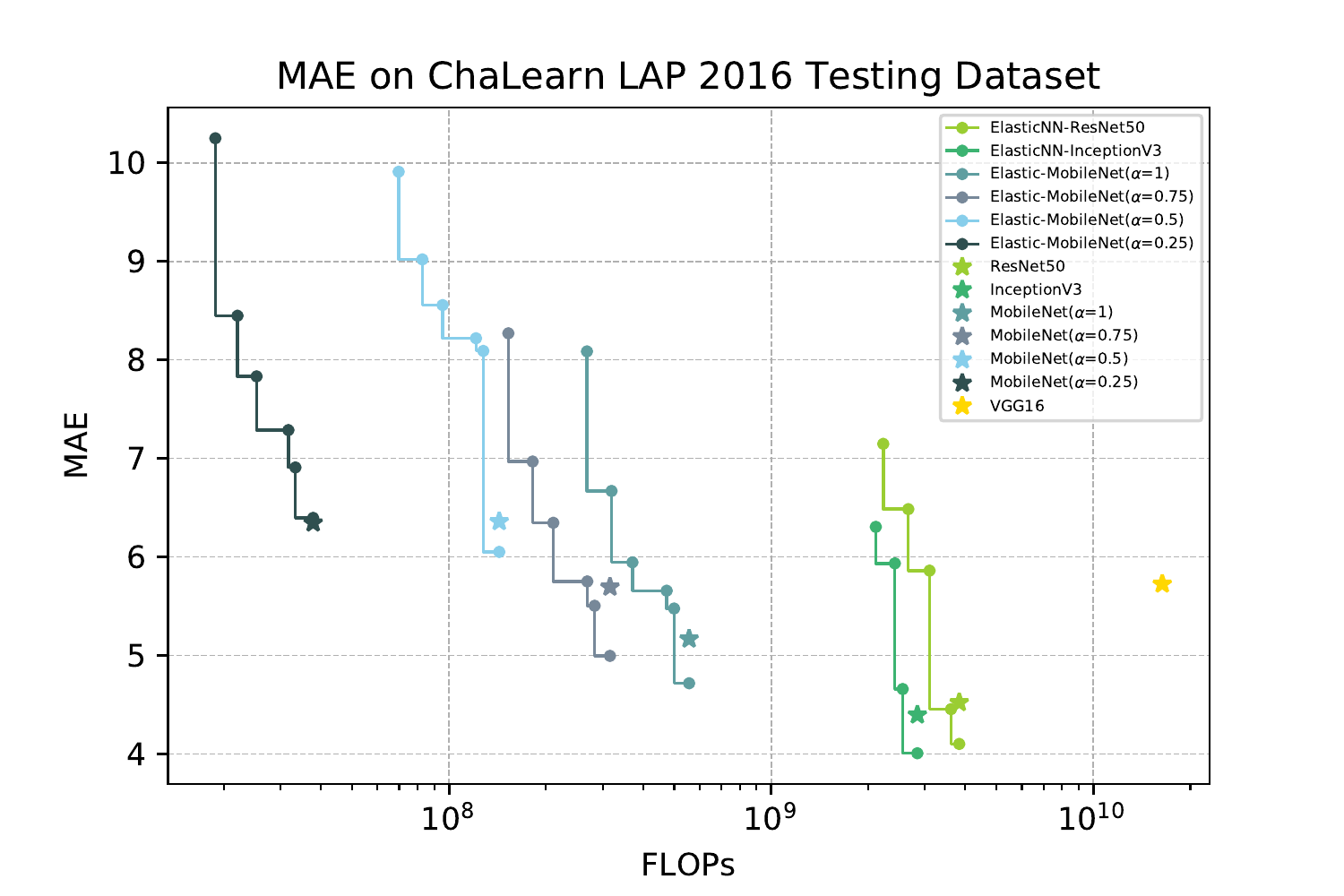} }}%
    \caption{Performance on different Elastic Framework based networks. Lower is better. }%
    \label{fig:result}%
\end{figure*}

Figure \ref{fig:result} shows the performance of the proposed elastic methodology implemented based on different networks on the ChaLearn LAP 2016 apparent age estimation testing dataset.

In particular, our elastic structures consistently exceed the performance of the corresponding original network structures. Table \ref{tab:elastic} displays the result at the situation with and without elastic structure separately. 
It shows that both MAE and $\epsilon$-error decrease notably for InceptionV3, ResNet50, and MobileNet ($\alpha = 1 / 0.75 / 0.5$, the $\alpha$ parameter in MobileNet increases or decreases the number of filters in each layer) on elastic structures. This indicates an interesting result that by introducing intermediate outputs act as regularizers, which benefits the final layer result.

Among elastic neural network architectures we carry out, InceptionV3 based network achieves the best performance with MAE and $\epsilon$-error equal to 4.01 and 0.3279 separately. Comparing to the result of the ChaLearn LAP 2016, our result has reached the third place. Notably, We only use a single network instead of combining the strength of different networks, besides we use a rather fast image preprocessing method which are more suitable for real-time tasks.

Table \ref{tab:flops} compares elastic neural network architectures in terms of their FLOPs requirements. It shows the FLOPs on the first intermediate output and on the last output separately of every elastic neural network architectures. The comparison includes the decreasing rate of FLOPs against the VGG16 \cite{simonyan2014very} neural network ($\Delta$ VGG) on the first and the last exit. The FLOPs on our proposed architectures decrease dramatically compare to VGG16, which is considered as an important benchmark for age estimation.

\begin{table}[t]
\vspace{3mm}
	\label{T:equipos}
	\caption{Test set accuracies of the ChaLearn LAP 2016 apparent age estimation dataset. The best $\epsilon$-error of each row in bold.}\label{tab.age_art}
	\begin{center}
    	\begin{threeparttable}
			\begin{tabular}{ l || c | c || c | c }
				\hline\hline
				 \multirow{2}{4em}{\textbf{Model}}& \multicolumn{2}{ c ||}{w/ Elastic Structure} & \multicolumn{2}{ c }{w/o Elastic Structure} \\ 
				\cline{2-5}
				& MAE & $\epsilon$-error & MAE & $\epsilon$-error \\
				\hline
                InceptionV3	&4.0070 & \textbf{0.3279}  &4.3967 & 0.3592	\\
                ResNet50    &4.1016 & \textbf{0.3428}	&4.5221 & 0.3682 \\
                MobileNet ($\alpha=1$)    &4.7174 & \textbf{0.3834}	&5.1672 & 0.4097\\
                MobileNet ($\alpha=0.75$)    &4.9952 & \textbf{0.4022}	& 5.6941 & 0.4311\\
                MobileNet ($\alpha=0.5$)    &6.0504 & \textbf{0.4296}	& 6.3596 & 0.4633\\
                MobileNet ($\alpha=0.25$)    &6.3953 & \textbf{0.4518}	& 6.3452 & 0.4753\\
                \hline
                VGG16 & - & - &5.7225 & 0.4338	\\
                CVPR2016 winner \textsuperscript{1} & - & - & - & 0.2411 \\
                CVPR2016 2nd \textsuperscript{1} & - & - & - & 0.3214 \\
                CVPR2016 3rd \textsuperscript{1} & - & - & - & 0.3361 
			\end{tabular}
            \begin{tablenotes}\footnotesize
                \item[1]  CVPR winners use more complicated preprocessing and/or multiple VGG16 style networks \cite{7789583}
			\end{tablenotes}
		\end{threeparttable}
	\end{center}\vspace{-0.5cm}
    \label{tab:elastic}
\end{table}
\begin{table}[t]
	\label{T:equipos}
	\caption{FLOPs of Elastic Framework for the networks compare to the state of art VGG16 network ($\Delta$ VGG).}\label{tab.age_art}
	\begin{center}
    \resizebox{\columnwidth}{!}{%
			\begin{tabular}{ l || c | c | c | c }
				\hline
				 \multirow{2}{4em}{\textbf{Model}}& \multicolumn{2}{ c |}{FLOPs} & \multicolumn{2}{ c }{$\Delta$ VGG} \\ 
				\cline{2-5}
				& first-exit & last-exit & first-exit & last-exit\\
				\hline
                VGG16 (w/o Elastic)	& \multicolumn{2}{c|}{$1.53 \times 10^{10}$ }&\multicolumn{2}{c}{0\%}	\\
                \hline
                InceptionV3	&$1.87\times 10^{9}$ &$2.84 \times 10^{9}$  & 87.83\% & 81.48\%\\
                ResNet50    &$2.23 \times 10^{9}$ & $3.83 \times 10^{9}$	&85.49\%  &75.02\%\\
                MobileNet ($\alpha=1$)    &$2.68 \times 10^{8}$ & $5.56 \times 10^{8}$	& 98.25\% & 96.38\% \\
                MobileNet ($\alpha=0.75$)    &$1.53 \times 10^{8}$ & $3.16 \times 10^{8}$	& 99\% & 97.94\% \\
                MobileNet ($\alpha=0.5$)    &$6.97 \times 10^{7}$ & $1.43 \times 10^{8}$	& 99.55\% & 99.07\% \\
                MobileNet ($\alpha=0.25$)    &\textbf{$1.88 \times 10^{7}$} & $3.78 \times 10^{7}$ 	& 99.88\% & 99.75\%\\
			\end{tabular}
		}
	\end{center}\vspace{-0.5cm}
    \label{tab:flops}
\end{table}







\section{Discussion and Conclusion}
\label{sec:conclusion}

We propose a general framework that systematically "elastifies" an arbitrary network to provide novel trade-offs between execution time and accuracy. This "elastification" is carried out by adding the so-called early-exits or intermediate outputs to any network topology. This framework enables us to design networks with adjustable computational load so that the network structure can elastically adapt to the load on the data stream rate basis. Choosing the intermediate outputs appropriately, we can improve both the computational speed and prediction accuracy compared to full network training, according to our experiments.

In our analysis, we made an interesting finding on the regularization effect of the so-called early exits in different network topologies. The early exits improve network generalization and is strongest with large networks having large expression power. This may open interesting future lines of research in studying if the effect applies in other network structures and applications, as well. 

As a use-case, we demonstrate the performance of our proposed framework with various commonly used pretrained deep networks for apparent age estimation. Our results show that the proposed apparent age estimator performs almost equally in comparison with recently reported apparent age estimators, but with significantly less computation. 

{
\bibliographystyle{IEEEtran}
\bibliography{IEEEabrv,ElasticNN.bib}
}

\end{document}